\documentclass[letterpaper, 10 pt, journal, twoside]{ieeetran} 
\usepackage{ifpdf}
\usepackage{algorithmic}
\usepackage{fixltx2e}
\usepackage[T1]{fontenc}
\usepackage[utf8]{inputenc}
\usepackage{cite}
\usepackage{amsmath,amsfonts}
\usepackage{amssymb}
\usepackage{array}
\usepackage{textcomp}
\usepackage{url}
\usepackage{verbatim}
\usepackage{graphicx}
\usepackage{graphics} 
\usepackage{mathtools}
\usepackage{booktabs}
\usepackage{adjustbox}
\usepackage{stfloats}
\usepackage[breaklinks=true,colorlinks,citecolor=blue]{hyperref}
\usepackage{rotating}
\usepackage{subfloat}
\usepackage{makecell}
\usepackage{multirow}
\usepackage{balance}
\usepackage{academicons}
\usepackage{xcolor}
\usepackage{lipsum}
\usepackage{colortbl}
\usepackage{rotating}
\usepackage[linesnumbered,ruled,vlined,commentsnumbered]{algorithm2e}
\usepackage[caption=false, font=footnotesize]{subfig}
\usepackage{footnote}
\usepackage{tabularray}
\usepackage{tablefootnote}
\usepackage[normalem]{ulem}
\UseTblrLibrary{siunitx}
\usepackage{amssymb}
\usepackage{pifont}

\begin{document}

\title{A Minimal Subset Approach for Informed\\ Keyframe Sampling in Large-Scale SLAM}

\makeatletter
\newcommand*\titleheader[1]{\gdef\@titleheader{#1}}
\AtBeginDocument{%
  \let\st@red@title\@title
  \def\@title{%
    \vskip-2.0em
    \bgroup\normalfont\small\centering\@titleheader\par\egroup
    \vskip1.5em\st@red@title}
}
\makeatother

\titleheader{Please cite as: N. Stathoulopoulos, C. Kanellakis and G. Nikolakopoulos, "A Minimal Subset Approach for Informed Keyframe Sampling in Large-Scale SLAM," in IEEE Robotics and Automation Letters, vol. 11, no. 1, pp. 738-745, Jan. 2026, doi: 10.1109/LRA.2025.3636035.}

%
%


\author{Nikolaos Stathoulopoulos, Christoforos Kanellakis and George Nikolakopoulos%
\thanks{This paper was supported in part and received funding from the European Union’s Horizon Europe Research and Innovation Programme under the Grant Agreement No.101138330.} 
\thanks{The authors are with the Robotics and AI Group, Department of Computer, Electrical and Space Engineering, Lule\r{a} University of Technology, 971 87 Lule\r{a}, Sweden.
    {Corresponding Author's e-mail: \tt\footnotesize niksta@ltu.se}}%
\thanks{Additional demonstration material and the code implementation of this letter are available at: \texttt{\href{https://github.com/LTU-RAI/opt-key}{https://github.com/LTU-RAI/opt-key.git}}.}
}



\maketitle
\begin{abstract}
Typical LiDAR SLAM architectures feature a front-end for odometry estimation and a back-end for refining and optimizing the trajectory and map, commonly through loop closures. However, loop closure detection in large-scale missions presents significant computational challenges due to the need to identify, verify, and process numerous candidate pairs for pose graph optimization.
Keyframe sampling bridges the front-end and back-end by selecting frames for storing and processing during global optimization. This article proposes an online keyframe sampling approach that constructs the pose graph using the most impactful keyframes for loop closure.
We introduce the Minimal Subset Approach (MSA), which optimizes two key objectives: redundancy minimization and information preservation, implemented within a sliding window framework. By operating in the feature space rather than 3-D space, MSA efficiently reduces redundant keyframes while retaining essential information.
Evaluations on diverse public datasets show that the proposed approach outperforms naive methods in reducing false positive rates in place recognition, while delivering superior ATE and RPE in metric localization, without the need for manual parameter tuning. Additionally, MSA demonstrates efficiency and scalability by reducing memory usage and computational overhead during loop closure detection and pose graph optimization.
\end{abstract}
\begin{IEEEkeywords}
SLAM, Place Recognition, Loop Closures
\end{IEEEkeywords}

\section{Introduction} \label{sec:introduction}
\IEEEPARstart{S}{imultaneous} Localization and Mapping (SLAM) in large-scale missions presents significant challenges, particularly in maintaining accuracy and efficiency as the environment grows over time. A key component of back-end SLAM is Pose Graph Optimization (PGO), which ensures consistent mapping by optimizing the global trajectory.
\begin{figure}[!t]
    \centering
    \includegraphics[width=0.9\columnwidth]{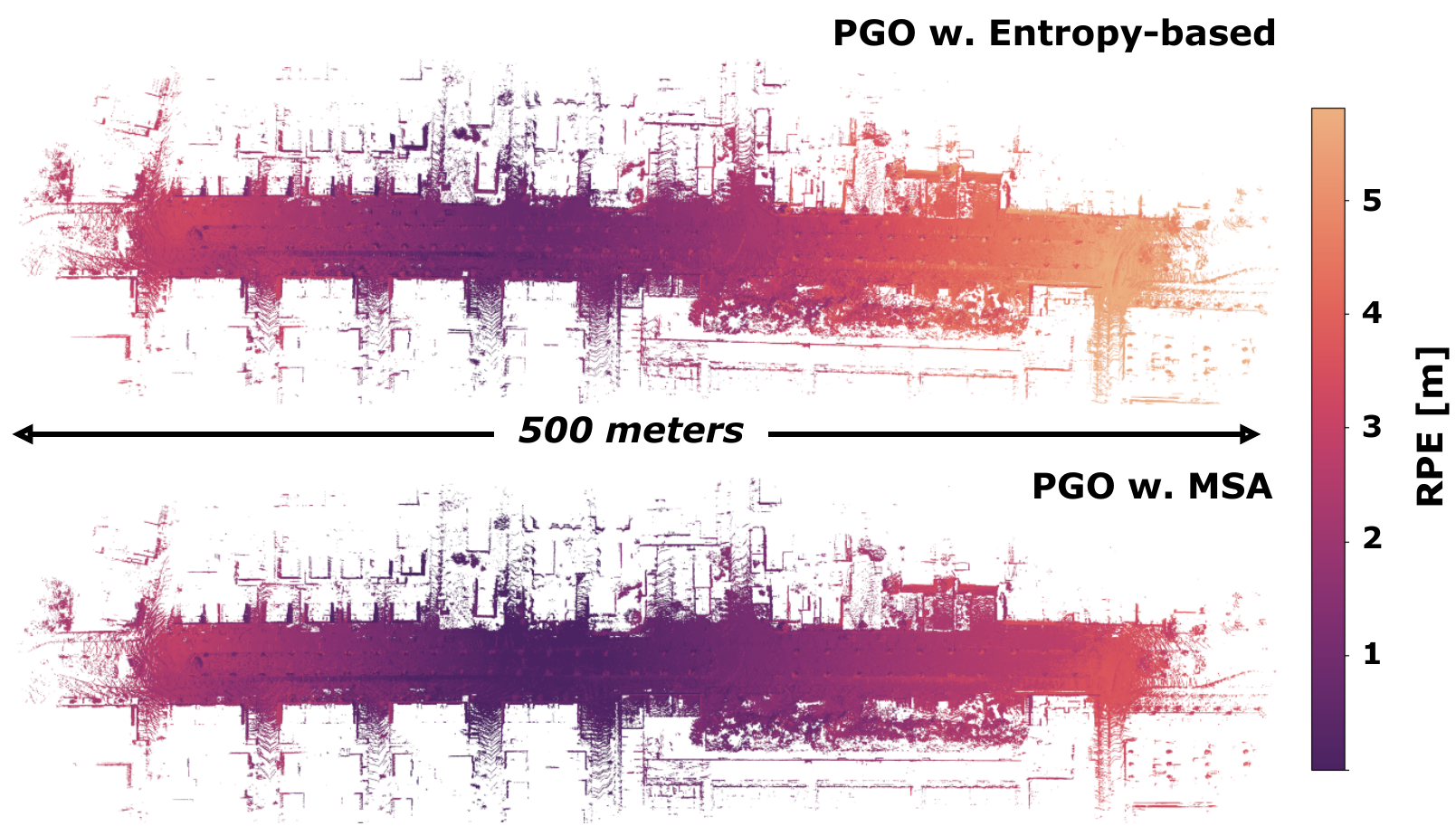}
    \caption{\textbf{An example on \texttt{KITTI 06}.} Comparison of two trajectories from \texttt{KISS-ICP}, after Pose Graph Optimization (PGO). In the top figure, keyframes are sampled using an entropy-based approach, while in the bottom figure, keyframes are sampled with the proposed Minimal Subset Approach (MSA) which achieves a lower Relative Pose Error (RPE).}
    \label{fig:concept}
    \vspace{-0.6cm}
\end{figure}
However, this process relies heavily on Loop Closure Detection (LCD) to reduce accumulated drift. In large-scale environments it becomes computationally expensive, as vast numbers of candidate pairs must be identified, verified, and processed to establish pose graph connections.
As the mission progresses, the number of loop closure candidates grows rapidly, leading to significant delays in processing times. Some datasets generate over 150,000 candidates, requiring four and a half hours to process, exceeding that of the mission time~\cite{Denniston2022LoopCP}. This complexity is further compounded by the computational intensity of data association and sequence matching during map optimization, where the processing time can range from seconds to several minutes, depending on the size of the data~\cite{stathoulopoulos2024frame}. 
Due to these computational challenges, many SLAM systems in large-scale environments have opted to either avoid or minimize loop closure detection to reduce the burden on limited computational resources. Some approaches, particularly in expansive environments, forego loop closures entirely, as the growing number of candidates becomes increasingly difficult to handle in real-time applications, as highlighted recently by Ebadi et al.~\cite{Kamak2024darpa_slam}. While loop closure detection is critical for correcting drift and ensuring map consistency, the computational complexity can overwhelm systems if not managed effectively, often leading to false positives that degrade the overall quality of the map~\cite{cadena2016slam}.

In order to address these challenges, our work centers on sampling keyframes online to reduce memory allocation and loop closure query times while preserving retrieval performance during Place Recognition (PR) and maintaining localization accuracy after pose graph optimization. We evaluate the impact of our optimized keyframe sampling method on critical back-end tasks, including loop closure detection and pose graph optimization, with a glimpse of the results showing in Fig.~\ref{fig:concept}. The proposed Minimal Subset Approach (MSA) employs two key criteria, \textit{redundancy minimization} and \textit{information preservation}, within a sliding window optimization framework to streamline keyframe selection. By reducing redundant information while preserving essential data, MSA enhances scalability and computational efficiency for both place recognition and pose graph optimization. 
Unlike traditional methods that rely on spatial heuristics in 3-D space, MSA operates in descriptor feature space. By leveraging the same extraction framework used for loop closure detection, it eliminates manual parameter tuning and achieves consistent performance across diverse scenarios.

Based on the aforementioned, the main contributions of this article can be summarized as follows:
(a) To the best of our knowledge, this is the first paper to present a keyframe sampling methodology that jointly evaluates the impact of sampling on place recognition performance and on the localization accuracy of the resulting pose graph optimization. We emphasize addressing these components together and discuss the theoretical and practical challenges that arise from this coupled view.
(b) We propose the Minimal Subset Approach (MSA), a keyframe sampling strategy that balances information preservation and redundancy minimization through a sliding window combinatorial optimization. By introducing principled metrics to quantify descriptor-space redundancy and information content, MSA selects a minimal set of keyframes that maintains loop closure potential and pose graph consistency, while remaining suitable for real-time application.
(c) Through our results, we demonstrate that MSA provides higher localization accuracy in terms of absolute trajectory error and stronger place  recognition performance across diverse environments, without manual parameter tuning. Compared to adaptive baselines such as entropy-based and spaciousness, MSA achieves state-of-the-art accuracy with comparable computational and memory demands, and compared to keeping all samples, it reduces loop closure query load and memory usage, making it well-suited for large-scale missions.
\vspace{-0.2cm}

\section{Related Work}

Efficient implementations of pose graph optimization 
have significantly enhanced its scalability by leveraging factor graphs and incremental optimization techniques. 
In more detail, \texttt{GTSAM}~\cite{gtsam} models probabilistic relationships using factor graphs and solves large-scale problems efficiently with solvers like \texttt{iSAM2}~\cite{Kaess2011isam2}, which incrementally updates the SLAM solution in real-time through a Bayes tree representation. Similarly, \texttt{g2o}~\cite{g2o} provides a versatile graph optimization framework, leveraging sparse pose graph connectivity for efficiency while supporting various parameterizations and loss functions.
Furthermore, sparsification methods~\cite{bai2021sparsification, Nam2024spectral} and hierarchical approaches~\cite{Guadagnino2022hipe, Tazaki2022multires} have further improved efficiency by reducing the number of active nodes in the optimization process.
Complementary to these back-end advances, Latif et al.~\cite{rrr2013} proposed \texttt{RRR}, which monitors pose-graph consistency over time and incrementally suppresses or removes spurious loop closures, improving robustness in long-term and multi-session mapping.
However, despite these advances, loop closure detection remains a major bottleneck due to the large number of candidate pairs that must be checked and processed to add meaningful constraints to the pose graph.

Keyframe sampling helps reduce the computational demands of loop closure detection by limiting the number of frames stored and processed in SLAM pipelines. In LiDAR-(Inertial) Odometry, systems such as \texttt{LIO-SAM}~\cite{liosam2020shan} and \texttt{DiSCo-SLAM}~\cite{Huang2022disco} or back-end global optimization systems like \texttt{LAMP}~\cite{Chang2022lamp2}, use fixed Euclidean distance intervals (1–2 meters) for keyframe generation to maintain local maps and optimize global poses. Although effective, these methods lack adaptability in dynamic environments with varying densities of loop closure candidates.
Recent advancements have introduced adaptive keyframe sampling techniques, adjusting intervals based on environmental spaciousness~\cite{chen2022dlo, Kim2023adaptive}, although manual threshold adjustment is still required. Entropy-based methods, like the one proposed by Zeng et al.~\cite{Zeng2023entropy}, use information theory to select keyframes, but face similar adaptability issues. Other approaches, such as the one proposed by Ou et al.~\cite{Ou2021}, focus on displacement vector similarity to balance computational cost and map completeness, while 
Chen et al.~\cite{degkey} select keyframes via online image-quality scoring (e.g., exposure, blur, occlusion), skipping degraded frames to improve visual localization.
In summary, while prior works have made progress in improving pose graph optimization scalability and in proposing various keyframe sampling heuristics, they often treat loop closure detection, place recognition, and keyframe selection as separate problems. Our approach differs by jointly formulating these components within a unified framework, ensuring that sampling decisions directly account for their downstream impact on both loop closure detection and pose graph optimization.

\section{Problem Formulation} \label{sec:problem_formulation}

\subsection{Pose Graph Optimization} \label{subsec:Pose graph optimization}
The objective of PGO is to find the most likely configuration of a robot's trajectory by minimizing the error in a graph-based representation of the poses. The pose graph is represented as a set of nodes and edges, where each node represents a robot pose and each edge encodes a spatial constraint between two poses~\cite{Grisetti2010graphslam}.
Let $\mathbf{X} = \{ \mathbf{x}_1, \ldots , \mathbf{x}_N\}$ be the set of $N$ robot poses, where, $\mathbf{x} \in SE(3)$, $\mathbf{z}_{ij} = h(\mathbf{x}_i, \mathbf{x}_j)$ be the observed spatial constraint between a pair of poses $\langle i, j \rangle$ and $\hat{\mathbf{z}}_{ij}(\mathbf{x}_i, \mathbf{x}_j)$ be the predicted value of that constraint.
The error $\mathbf{e}_{ij} \in \mathbb{R}^6$ quantifies the difference between the observed relative transformation and the predicted transformation between poses $\mathbf{x}_i$ and $\mathbf{x}_j$ and is denoted as:
\begin{equation} \label{eq:error}
    \mathbf{e}_{ij} = \mathbf{e}(\mathbf{x}_i, \mathbf{x}_j) = \mathbf{z}_{ij} - \hat{\mathbf{z}}_{ij}
\end{equation}
We can then define the set of edges $\mathcal{E} = \big\{ \langle i, j \rangle \: | \: \exists \, \mathbf{z}_{ij}\big\}$, that contains all the pairs in which a relative constraint exists. The goal of pose graph optimization is to find the optimal configuration of poses $\mathbf{X}^*$ that minimizes the overall error function $E(\mathbf{X})$, denoted as:
\begin{equation} \label{eq:Pose graph optimization}
\mathbf{X}^* = \operatorname*{arg\,min}_{\mathbf{X}} \sum_{\langle i, j \rangle \in \mathcal{E}} \mathbf{e}(\mathbf{x}_i, \mathbf{x}_j)^{\top} \, \mathbf{\Omega}_{ij} \, \mathbf{e}(\mathbf{x}_i, \mathbf{x}_j),
\end{equation}
where $\mathbf{\Omega}_{ij} \in \mathbb{R}^{6 \times 6}$ is the information matrix that reflects the certainty of the relative pose measurement for each pair.

\subsection{Odometry and Loop Closure Edges}
Pose graph optimization is widely used as a global optimization method to mitigate accumulated pose drift during large-scale and long-term missions. This would not be possible without loop closures. Relying solely on pose-to-pose odometry constraints is insufficient to maintain global consistency, hence, loop closures are essential. 
Given this, we define two distinct subsets in the set of edges $\mathcal{E}$ in the pose graph.
The first subset, $\mathcal{E}_{o} \subseteq \mathcal{E}$, consists of \textit{odometry edges}, which represent consecutive movements of the robot based on local motion estimates, while the second subset, $\mathcal{E}_{l} \subseteq \mathcal{E}$, consists of \textit{loop closure edges}, which correspond to observations of previously visited locations.
Thus, the overall error function can be decomposed into two components:
\begin{equation}
    E(\mathcal{\mathbf{X}}) = \sum_{\langle i, j \rangle \in \mathcal{E}_{o}} \mathbf{e}_{ij}^{\top} \mathbf{\Omega}_{ij} \, \mathbf{e}_{ij} + \sum_{\langle i, j \rangle \in \mathcal{E}_{l}} \mathbf{e}_{ij}^{\top} \mathbf{\Omega}_{ij} \, \mathbf{e}_{ij},
\end{equation}
where the edge sets satisfy the following two relationships: $\mathcal{E} = \mathcal{E}{o} \cup \mathcal{E}_{l}$ and $\mathcal{E}_{o} \cap \mathcal{E}_{l} = \varnothing$.
\vspace{-0.2cm}

\subsection{Loop Closure Detection} \label{subsec:lcd}
Detecting loop closures is challenging, particularly in large-scale missions, where accumulated drift can cause traditional radius-based searches to fail by missing nearby candidates. A common solution is \textit{place recognition}, which compares current sensory data (e.g., images or LiDAR scans) against a database of past observations to identify matches. To formalize this, we define a keyframe set $\mathbf{K} = \{ (\mathbf{x}_i, \mathbf{d}_i, \mathbf{z}_i) \, | \, i = 1, \ldots, N \}$, where $\mathbf{x}_i \in SE(3)$ is the pose of keyframe $i$, $\mathbf{d}_i \in \mathbb{R}^M$ represents its descriptor which encodes distinct features~\cite{ma2022overlaptrans, Giseop2022scpp}, and $\mathbf{z}_i$ is the set of associated measurements. The goal is to find loop closure candidates $\mathcal{L}$ by comparing descriptors:
\begin{equation} \label{eq:loops}
    \mathcal{L} = \big\{ \langle i, j\rangle \mid f(\mathbf{d}_i, \mathbf{d}_j) > \tau, \; \forall \langle \mathbf{k}_i, \mathbf{k}_j \rangle \in \mathbf{K} \times \mathbf{K}, \; \mathbf{k}_i \neq \mathbf{k}_j \big\},
\end{equation}
where $f(\mathbf{d}_i, \mathbf{d}_j)$ is a similarity function (e.g., cosine similarity, Euclidean distance) and $\tau$ is a threshold for considering two keyframes representing the same place. This approach generalizes to various scenarios, such as comparing a current keyframe with the $k$ nearest past keyframes in single-robot detection or comparing keyframes across multiple robots.

\subsection{Reducing Search Space in Place Recognition} \label{subsec:pr}
The detection and verification of loop closure edges $\mathcal{E}_l \triangleq \mathcal{L}$ is computationally demanding due to the large number of potential keyframe pairs, which scales quadratically as $O(n^2)$, where $n = |\mathbf{K}|$ is the number of keyframes. While using descriptors and methods like \textit{k}-nearest neighbor (\textit{k}-NN) search~\cite{yin2023survey} can accelerate the process, additional steps are still required to remove outliers, mitigate false positives, or prioritize certain nodes~\cite{Denniston2022LoopCP}, while extracting the relative pose transformation between loop closure nodes also involves computationally intensive techniques such as General Iterative Closest Point (GICP)~\cite{GICP}. This pipeline does not scale well for extensive missions, often resulting in processing times of 3-4 hours or more~\cite{Kamak2024darpa_slam, Chang2022lamp2}, and the challenge becomes even greater in multi-robot systems~\cite{stathoulopoulos2024frame}. 

To address this computational complexity, we propose a \textit{sampling strategy} to reduce the number of keyframes considered for matching, while maintaining minimal impact on the loop closure set $\mathcal{L}$. Let $\mathbf{K}_\text{S} \subset \mathbf{K}$ represent a subset of keyframes selected through a sampling process $S: \mathbf{K} \rightarrow \mathbf{K}_\text{S}$ such that $|\mathbf{K}_\text{S}| < |\mathbf{K}|$. Our objective is to reduce the search space by confining the loop closure detection process to only the selected keyframes in $\mathbf{K}_\text{S}$. The goal is to identify a reduced set $\mathbf{K}_\text{S}^*$ that minimizes redundant keyframes while preserving the set of loop closures $\mathcal{L}$, a process denoted as:
\begin{equation} \label{eq:minimum-cardinality}
    \mathbf{K}_\text{S}^* = \operatorname*{arg\,min}_{\mathbf{K}_\text{S} \subseteq \mathbf{K}} \left( |\mathbf{K}_\text{S}| \right), \; \text{subject to} \; \mathcal{L}^* = \mathcal{L}, \text{ where }
\end{equation}
\begin{equation} \label{eq:cosnt_loops}
    \mathcal{L}^* = \big\{ \langle i, j\rangle \mid f(\mathbf{d}_i, \mathbf{d}_j) > \tau, \; \forall \langle \mathbf{k}_i, \mathbf{k}_j \rangle \in \mathbf{K}^*_\text{S} \times \mathbf{K}^*_\text{S}\big\},
\end{equation}
In this approach, the overall pose graph optimization is reformulated as follows:
\begin{equation} \label{eq:pgo_card}
    \mathbf{X}^*_\text{S} = \operatorname*{arg\,min}_{\mathbf{X}_\text{S}} \sum_{\langle i, j \rangle \in \mathcal{E}^*_o} \mathbf{e}_{ij}^{\top} \, \mathbf{\Omega}_{ij} \, \mathbf{e}_{ij} + \sum_{\langle i, j \rangle \in \mathcal{E}_l^*} \mathbf{e}_{ij}^{\top} \, \mathbf{\Omega}_{ij} \, \mathbf{e}_{ij},
\end{equation}
where $\mathbf{X}_\text{S}$ represents the poses corresponding to the optimally sampled keyframe set $\mathbf{K}_\text{S}^*$, and $\mathcal{E}_o^*,\,\mathcal{E}_l^* \triangleq \mathcal{L}^*$ denote the odometry and loop closure edge pairs between the sampled keyframe poses $\mathbf{X}_\text{S}$, respectively.
\vspace{-0.2cm}

\subsection{Underlying Challenges}
Addressing the combined problem of pose graph optimization and loop closure detection through place recognition, presents several underlying challenges that affect both the efficiency and accuracy of the solution.
\subsubsection{Impact of Sampling} 
\begin{figure}[t!]
    \centering
    \includegraphics[width=0.85\columnwidth]{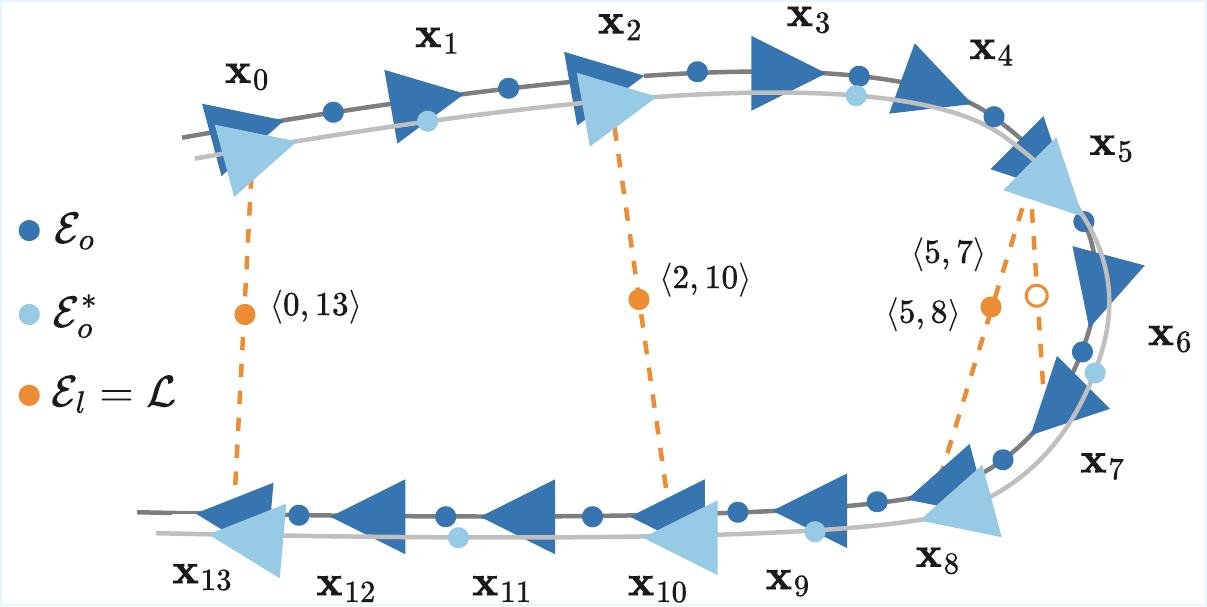}
    \vspace{-0.15cm}
    \caption{\textbf{Redundant keyframes effect.} Example of removing redundant keyframes (light blue). The updated odometry edge set, $\mathcal{E}_o^*$, influences the pose graph optimization, while the sparsified loop closure edges, $\bar{\mathcal{L}} = \mathcal{L} \:\backslash \langle 5,7 \rangle$, can further reduce the computational complexity.}
    \label{fig:redundancy}
    \vspace{-0.6cm}
\end{figure}

The sampling strategy $S$ reduces $\mathbf{K}$ to a smaller subset $\mathbf{K}_\text{S}$ to keep place recognition scalable, thereby altering the loop closure set and the odometry edges. Over-aggressive pruning can remove informative loop closures needed to correct drift, and over-coarsen the odometry chain, increasing linearization error and degrading PGO conditioning. It may also leave regions under-represented in descriptor space, weakening place recognition retrieval on revisit. Conversely, keeping many near-duplicate frames bloats place recognition search and verification, increases false positives in self-similar areas, and on the PGO side, over-weights correlated constraints, slowing or destabilizing the solver. The core challenge is to select a subset that preserves the essential loop closure structure while maintaining well-spaced consecutive poses, as depicted in Fig.~\ref{fig:redundancy}.
\subsubsection{Combined Problem} 
Place recognition and pose graph optimization solve different subproblems. Place recognition operates in high-dimensional descriptor space, ranking appearance-based matches and remaining sensitive to viewpoint and self-similar structure; its errors show up as false positives or missed matches. PGO operates in geometric space on poses in $SE(3)$, solving a sparse nonlinear least-squares problem whose behavior depends on graph connectivity, scaling, and numerical conditioning; failures arise from inconsistent or poorly scaled constraints and weak observability. The challenge is to reconcile appearance-based data association with metric consistency over the pose graph.
\subsubsection{Dynamic Keyframe Selection} 
To minimize the keyframe set size in real-time, we must consider the entire set $\mathbf{K}$; however, this is impractical for maintaining computational feasibility in loop closure detection. The challenge is to determine the contribution of each keyframe, which requires anticipating future keyframes and introduces a lack of causality. Moreover, finding the optimal subset $\mathbf{K}_\text{S}^*$ is an NP-hard problem, involving an exhaustive search of all keyframe combinations, which becomes computationally infeasible for sets larger than 15--20 keyframes.
The goal of this research is to show that reducing the number of keyframes through an informed sampling strategy can significantly decrease the search space for loop closure detection, improving computational efficiency by reducing memory usage and query time without compromising SLAM accuracy. The proposed approach seeks to balance computational complexity and optimization precision, ensuring scalability and robustness.
\vspace{-0.2cm}

\section{Minimal Subset Approach} \label{sec:otpimization}
To address these challenges, we propose the Minimal Subset Approach (MSA) to approximate a solution to the problems in Eq.~(\ref{eq:minimum-cardinality})-(\ref{eq:cosnt_loops}). 
The proposed MSA identifies and eliminates redundancy within a keyframe set while preserving essential data. Designed for real-time use, it employs a sliding window combinatorial optimization with dual objectives: balancing \textit{information preservation} and~\textit{redundancy minimization}.
\vspace{-0.3cm}

\subsection{Redundancy in Keyframes} \label{subsec:rmt}
The frequency of a sensor, the platform's speed, and the environmental characteristics can cause keyframe samples to capture redundant information if they are too closely spaced.
To address this, we define redundancy within a keyframe set and propose a metric to quantify it based on the descriptor space. A keyframe $\mathbf{k}$ is redundant within a set $\mathbf{K}$ if its removal does not affect the loop closure edge set $\mathcal{L}$ and does not create discontinuities in the map representation:
\begin{equation*} \label{eq:loops}
     \bar{\mathcal{L}} \equiv \mathcal{L} = \big\{ \langle i, j\rangle \mid f(\mathbf{d}_i, \mathbf{d}_j) > \tau, \; \forall \langle \mathbf{k}_i, \mathbf{k}_j \rangle \in \bar{\mathbf{K}} \times \bar{\mathbf{K}}\big\},
\end{equation*}
\begin{equation} \label{eq:constrain}
    \operatorname*{subject\:to}\:\:\: \delta_{l} \leqslant \| \mathbf{x}_i - \mathbf{x}_{i+1} \|_2 \leqslant \delta_{u},\, \forall \, \mathbf{x} \in \mathbf{X},
\end{equation}
where $\bar{\mathbf{K}} = \mathbf{K} \,\backslash \{\mathbf{k}\}$. Here, “$\backslash$” denotes set difference: remove $\mathbf{k}$ from $\mathbf{K}$.
The spatial constraint in Eq.~(\ref{eq:constrain}) ensures comprehensive map coverage and that consecutive odometry edges can be computed safely. 
We quantify redundancy in a keyframe set using a metric that captures the similarity between consecutive keyframes, utilizing any similarity function $f_\sigma$, provided by the corresponding descriptor extraction framework:
\begin{equation} \label{eq:redundancy_term}
    \rho_{\tau}(\mathbf{K}) = \frac{1}{N} \sum_{i=0}^{N-1} f_{\sigma} (\mathbf{k}_i,\mathbf{k}_{i+1}),
\end{equation}
where $0 < \rho_{\tau}(\mathbf{K}) \leqslant 1$ and $N$ is the number of keyframes in $\mathbf{K}$. Higher values indicate greater redundancy.
\vspace{-0.15cm}

\subsection{Information Preservation in Keyframes} \label{subsec:ipt}
Let $F:\mathbb{R}^3 \rightarrow \mathbb{R}^M$, represent the descriptor extraction function, either learning-based~\cite{ma2022overlaptrans, cattaneo_lcdnet_2022} or handcrafted~\cite{Giseop2022scpp, Xu2023ring}, which maps each observation from 3-D space to an $M$-dimensional representation. This function does not directly depend on the pose $\mathbf{x}_t$, but rather on the input observation $\mathbf{z}_t$ (e.g. LiDAR scan), which itself depends on the pose through the sensor model $\mathbf{z}_t \propto h(\mathbf{x}_t,\mathbf{W})$, where $\mathbf{W}$ represents the world.
To understand how the descriptors are sensitive to pose changes, we can compute the Jacobian $\mathbf{J}$ of $F$ with respect to poses $\mathbf{x}$ through the chain rule $(\partial \mathbf{F} \, \partial \mathbf{z})/(\partial \mathbf{z} \, \partial \mathbf{x})$.
Explicitly deriving the Jacobian can be impractical because of the intricate nature of functions like deep neural networks. Consequently, we employ numerical approximations to estimate the rate of change. Moreover, since poses are encompassed within $SE(3)$, we reduce the dimensional complexity by using the Euclidean norm for distance measurement instead of individual axis derivatives. It is essential to use yaw-invariant descriptors, as suggested in \cite{ma2022overlaptrans, Giseop2022scpp}, to ensure the Jacobian calculations are meaningful by preventing orientation changes. 
Considering the descriptors $\mathbf{d} \in \mathbb{R}^{M}$ as random variables and poses $\mathbf{x} \in SE(3)$ as samples, the product $\mathbf{J}_{\mathbf{F}}^\mathsf{T} \mathbf{J}_{\mathbf{F}}$ estimates the information matrix: $\mathbf{J}_{\mathbf{F}}^\mathsf{T} \mathbf{J}_{\mathbf{F}} = ({\partial \mathbf{F}}/{ \partial \mathbf{x}})^\mathsf{T}({\partial \mathbf{F}}/{ \partial \mathbf{x}} ) = \mathbf{V}\mathbf{\Lambda}{\mathbf{V}^{-1}}$,
where $\mathbf{\Lambda}$ is the diagonal matrix of eigenvalues and $\mathbf{V}$ is the matrix of eigenvectors from the decomposition. The eigenvectors represent the principal directions of variation, and the eigenvalues represent their magnitudes. The descriptors are transformed using the eigenvectors and scaled by the square root of the eigenvalues, $\mathbf{D}' = \sqrt{\mathbf{\Lambda}} \cdot \mathbf{V} \cdot \mathbf{D}$, aligning them with the main directions of variability.
Therefore, we can define the \textit{information preservation} term for a keyframe set as:
\begin{equation} \label{eq:info_preserv}
    \pi_\tau(\mathbf{K}) = \frac{1}{N} \sum_{i=0}^{N-1} f_\delta(\mathbf{d}_{i}',\mathbf{d}_{i+1}'),
\end{equation}
where $0 < \pi_{\tau}(\mathbf{K}) \leqslant 1$, $f_\delta$ is the distance function between descriptors, and $\mathbf{d}' \in \mathbf{D}'$ are the transformed descriptors. Higher values of $\pi_\tau$ indicate that the poses better preserve the variability in the descriptor space.
Eigenvectors $\mathbf{v}_k$ denote directions of maximal variability, and the $k$-th eigenvalue $\lambda_k$ quantifies the variance described by each eigenvector. Larger eigenvalues indicate more significant patterns of variability. 

In summary, while both $\rho_\tau$ and $\pi_\tau$ utilize a similarity or distance function, they differ in objectives: the redundancy term focuses on local redundancy within the keyframe set, while the information preservation term evaluates the preservation of information structure with respect to pose changes. The distinct goals lead to different interpretations of relationships between descriptive vectors within the keyframe set.
\vspace{-0.2cm}

\subsection{Sliding Window Optimization} \label{subsec:sliding_window}
\begin{figure}[t!]
    \centering
    \setlength{\abovecaptionskip}{-11pt}
    \includegraphics[width=1.\columnwidth]{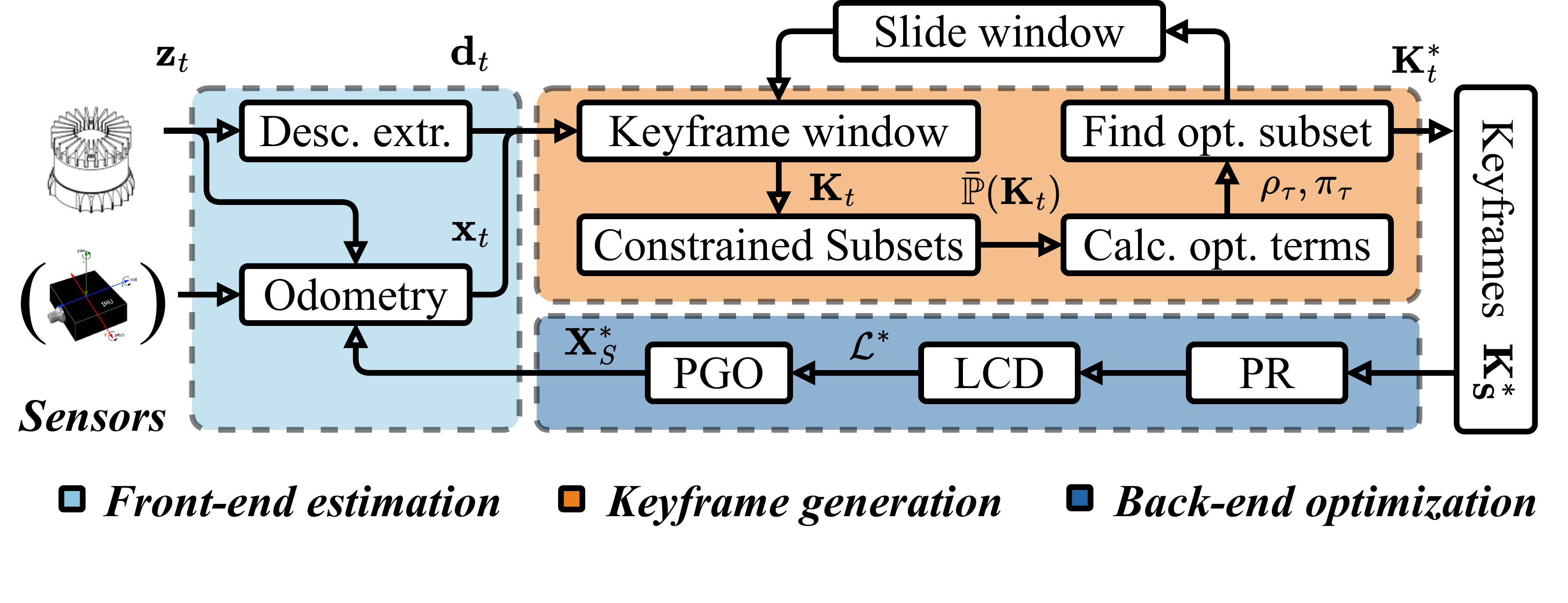}
    \caption{\textbf{Overall pipeline.} Block diagram of the overall pipeline, highlighting the interplay between front-end (descriptor extraction and odometry modules) and back-end optimization (place recognition, loop closure detection and pose graph optimization), facilitated by the proposed keyframe sampling scheme. }
    \label{fig:}
    \vspace{-0.65cm}
\end{figure}

To manage the computational complexity of optimizing a large keyframe set and proactively selecting keyframes for future queries, we introduce a sliding window optimization method. 
Figure~\ref{fig:} illustrates the interaction between the front-end and back-end systems, highlighting the contribution of the proposed keyframe sampling strategy in linking the two. 
This approach continuously optimizes keyframe sets $\mathbf{K}_t$ over the mission time $T$, approximating the optimal keyframe set as:

\begin{equation} \label{eq:approx_keyframe_set}
    \mathbf{K}_{\text{S}}^* \cong \bigcup_{t \, \in \, \left[0,T\right]}\mathbf{K}_t^*,
\end{equation}
where $\mathbf{K}_t^*$ is the minimal subset of $\mathbf{K}_t$ that retains maximum information. To compute this subset, we use the redundancy and information preservation terms defined earlier.

The process begins by initializing a window keyframe set $\mathbf{K}_t$ containing $N$ keyframes. The time step $t$ advances once $N$ new keyframes are available, and the optimization gets the optimal window keyframe set $\mathbf{K}_t^*$. 
The last selected keyframe becomes the first in the next window, and any previously unselected keyframes are re-evaluated in the context of the newly incoming keyframes.
To find this optimum, we generate all possible subsets of $\mathbf{K}_t$, forming the power set $\mathbb{P}(\mathbf{K}_t)$ with cardinality $|\mathbb{P}(\mathbf{K}_t)| = 2^N$. 
To reduce computational complexity, we apply constraints on the power set, retaining only subsets that satisfy min and max distances between consecutive poses, as per Eq.~(\ref{eq:constrain}), reducing the power set by 5–10 times. The constrained power set can be denoted as:
\begin{equation} \label{eq:power_set}
    \bar{\mathbb{P}}(\mathbf{K}_t) = \big\{\mathbf{K}_t^{\mathbb{S}} \in \mathbb{P}(\mathbf{K}_t)| \delta_{l} \leqslant \|\mathbf{x}_i - \mathbf{x}_{i+1}\|_2 \leqslant \delta_{u}, \forall \mathbf{k} \in \mathbf{K}_t^{\mathbb{S}}\big\},
\end{equation}
where $\delta_l$ and $\delta_u$ are the lower and upper bounds on inter-pose distance ($\delta_l$ prevents near-duplicate/stationary frames; $\delta_u$ keeps pairs within descriptor locality).
We adaptively select these limits based on the average frame-to-frame distance $\delta_\text{avg}$ within a window as $\delta_l = \lambda_l \cdot \delta_\text{avg}$ and $\delta_u = \lambda_u \cdot \delta_\text{avg}$, where $\lambda_l$ and $\lambda_u$ are constant hyperparameters, further discussed in Section~\ref{subsec:ablation}.
The window optimization can be formulated as:
\begin{align} \label{eq:objective}
    \mathbf{K}_t^* = &\operatorname*{arg\,min}_{\mathbf{K}_t^{\mathbb{S}}} \:\: \big({\rho_{\tau}\,(\mathbf{K}_t^{\mathbb{S}}) + \alpha }\big)\big/\big({\pi_\tau(\mathbf{K}_t^{\mathbb{S}})} + \beta \big) \\
    & \operatorname*{where} \:\:  \alpha,\,\beta > 0 \:\operatorname*{ and }\: \mathbf{K}_t^{\mathbb{S}} \in \bar{\mathbb{P}}(\mathbf{K}_t).
\end{align}
The minimization problem is solved through an exhaustive search, evaluating each subset's information matrix $\mathbf{J}_{\mathbf{F}}^\mathsf{T} \mathbf{J}_{\mathbf{F}}$ and quantifying the information preservation and redundancy terms, selecting the subset with the best combined score.

\section{Experimental Setup and Results}
\begin{table*}[!t]
\centering
\setlength{\abovecaptionskip}{-2pt}
\caption{Translational (t) and Rotational (R) Absolute Trajectory Error Improvement after PGO and the Area Under the PR Curve$^{*}$}
\label{tab:global_evaluation}
\resizebox{\linewidth}{!}{%
\begin{tabular}{c||l||c|c|c||c|c|c||c|c|c} 
\toprule
& \textbf{Datasets} & \multicolumn{3}{c||}{\textbf{\texttt{KITTI Odometry} (Seq. \texttt{05} / \texttt{06} )}} & \multicolumn{3}{c||}{\textbf{\texttt{MulRan} (\texttt{DCC} / \texttt{KAIST})}} & \multicolumn{3}{c}{\textbf{\texttt{Apollo-SB} (\texttt{SanJose} / \texttt{Columbia})}} \\ 
\midrule
& \textbf{Metrics} & \textbf{t. ATE [\%]} & \textbf{R. ATE [\%]} & \textbf{PR-AUC [\%]} & \textbf{t. ATE [\%]} & \textbf{R. ATE [\%]} & \textbf{PR-AUC [\%]} & \textbf{t. ATE [\%]} & \textbf{R. ATE [\%]} & \textbf{PR-AUC [\%]} \\ 
\midrule
\multirow{7}{*}{\rotatebox{90}{\textbf{\texttt{ScanContext}}}} & All Samples  & 72.7 / 76.1 & 94.4 / 49.0 & 99.8 / 99.4 & 30.2 / 69.4 & 24.1 / 56.3 & 99.6 / 99.1 & 92.0 / 93.7 & 93.7 / 92.3 & 74.6 / 91.0 \\
& Const. 1.0m & \textbf{90.5} / \underline{76.2} & \underline{95.9} / \underline{52.6} & \underline{97.1} / \textbf{96.6} & 63.2 / \underline{88.6} & 55.2 / 70.7 & \textbf{96.1} / \underline{85.9} & {84.3} / \textbf{92.5} & {87.4} / \textbf{92.0} & 62.5 / \textbf{89.8} \\
& Const. 2.0m & 86.2 / 62.3 & 94.0 / 46.4 & 71.8 / 78.2 & \textbf{72.1} / 83.3 & \underline{69.4} / \underline{72.0} & 79.8 / 83.4 & 66.7 / 79.2 & 59.5 / 78.6 & 51.2 / 81.3 \\ 
& Const. 3.0m & 79.8 / 60.1 & 77.4 / 42.3 & 74.2 / 74.5 & 58.7 / 79.5 & 60.6 / 69.2 & 64.2 / 73.5 & 54.4 / 77.1 & 40.9 / 69.8 & 41.0 / 68.2\\ 
& Entropy-based & 66.9 / 46.7 & 93.4 / 48.3 & 92.0 / 90.9 & 68.7 / 71.8 & 69.0 / 66.8 & 89.1 / 82.8 & \underline{90.1} / 80.3 & \underline{89.8} / {82.9} & \underline{72.5} / \underline{85.8} \\
& Spaciousness & 77.8 / 55.6 & 87.6 / 46.7 & 76.5 / 77.8 & 67.9 / 70.7 & 68.5 / 66.1 & 81.0 / \textbf{86.3} & 79.7 / 79.6 & 84.2 / 82.1 & 45.5 / 79.3 \\
& {MSA (Ours)} & \underline{90.3} / \textbf{76.4} & \textbf{96.0} / \textbf{53.2} & \textbf{99.7} / \underline{94.0} & \underline{71.9} / \textbf{91.8} & \textbf{70.0} / \textbf{76.8} & \underline{95.3} / {85.7} & \textbf{91.6} / \underline{91.5} & \textbf{91.4} / \underline{90.2} & \textbf{74.3} / {85.6} \\
\midrule
\multirow{7}{*}{\rotatebox{90}{\textbf{\texttt{OverlapTrans}}}} & All Samples   & 75.7 / 78.6  & 95.9 / 51.1 & 99.9 / 99.5 & 29.8 / 67.4 & 24.9 / 59.5 & 99.9 / 89.8 & 96.2 / 98.1 & 95.3 / 97.5 & 75.5 / 90.2  \\ 
& Const. 1.0m & \textbf{94.6} / \underline{78.7}  & \underline{97.3} / \underline{55.3} & \underline{97.3} / \textbf{96.7} & 59.3 / 87.1 & 56.4 / 71.0 & \textbf{96.6} / \textbf{86.7} & \underline{85.8} / \underline{97.1} & \underline{84.1} / \underline{97.2} &  {62.8} / \textbf{89.9} \\ 
& Const. 2.0m   & 84.4 / 44.9 & 97.2 / 23.7 & 72.1 / 78.0 & \underline{70.0} / 80.9 & {68.4} / 73.7 & 80.7 / 83.9 & 06.8 / 75.9 & 06.8 / 77.2 & 50.7 / 81.4 \\ 
& Const. 3.0m   & 90.5 / 22.6 & 87.9 / 13.8  & 74.3 / 74.2 & 63.6 / \underline{87.7} & 64.6 / \textbf{79.1} & 64.1 / 72.9 & 00.0 / 83.6 & 00.0 / 84.4 & 41.3 / 67.9 \\ 
& Entropy-based & 67.9 / 24.1 & 93.0 / 09.1 & 92.2 / 89.7 & 69.2 / 68.7 & \underline{69.5} / 63.1 & 89.4 / 83.8 & 77.6 / 69.4 & 76.2 / 72.6 & \underline{72.1} / \underline{86.0} \\ 
& Spaciousness  & 85.8 / 35.5 & 97.2 / 39.1 & 76.4 / 77.5 & 68.4 / 68.6 & 68.1 / 63.1 & 81.4 / 86.0 & 03.5 / 67.9 & 03.5 / 71.7 & 42.7 / 79.9 \\ 
& {MSA (Ours)}    & \underline{93.1} / \textbf{78.8} & \textbf{97.6} / \textbf{55.5} & \textbf{99.8} / \underline{93.8} & \textbf{70.2} / \textbf{91.2} & \textbf{69.9} / \underline{77.5} & \underline{95.5} / \underline{86.3} & \textbf{93.0} / \textbf{98.1} & \textbf{92.0} / \textbf{97.5} & \textbf{75.0} / {85.2} \\
\bottomrule
\multicolumn{11}{c}{\vspace{-0.25cm}} \\ 
\multicolumn{11}{c}{$^{*}$The \textbf{bold} results correspond to the best performing sampling method, while the \underline{underlined} results correspond to the second best  performing sampling method.}
\end{tabular}
}
\vspace{-0.4cm}
\end{table*}

To support the claims of our contributions, on how the sampling interval affects both place recognition and pose graph optimization performance, we conducted a series of experiments with the following setup. Descriptor extraction was performed using two approaches: the learning-based \texttt{OverlapTransformer} (\texttt{OT})~\cite{ma2022overlaptrans}, which generates a $1 \times 256$ feature vector, and the hand-crafted \texttt{Scan Context} (\texttt{SC})~\cite{Giseop2022scpp}, which produces a $20 \times 60$ feature representation. These choices demonstrate the descriptor-agnostic nature of our method and its compatibility with both learned and hand-crafted descriptors.
We benchmark our approach against constant sampling intervals of 1, 2, and 3 meters, commonly found in algorithms such as \texttt{LIO-SAM}~\cite{liosam2020shan}, \texttt{LAMP}~\cite{Chang2022lamp2} and \texttt{DiSCo-SLAM}~\cite{Huang2022disco}, as well as adaptive sampling strategies based on LiDAR spaciousness \cite{chen2022dlo} and entropy \cite{Zeng2023entropy}. All experiments were conducted on a 14th Gen Intel Core i9 system with 128GB of RAM, where the sampling window optimization ($N=10$) is solved in approximately 5–15 milliseconds, making it suitable for real-time sampling. 
\vspace{-0.2cm}

\begin{figure}[!t]
    \centering
    \setlength{\abovecaptionskip}{-2pt}
    \includegraphics[width=1.0\columnwidth]{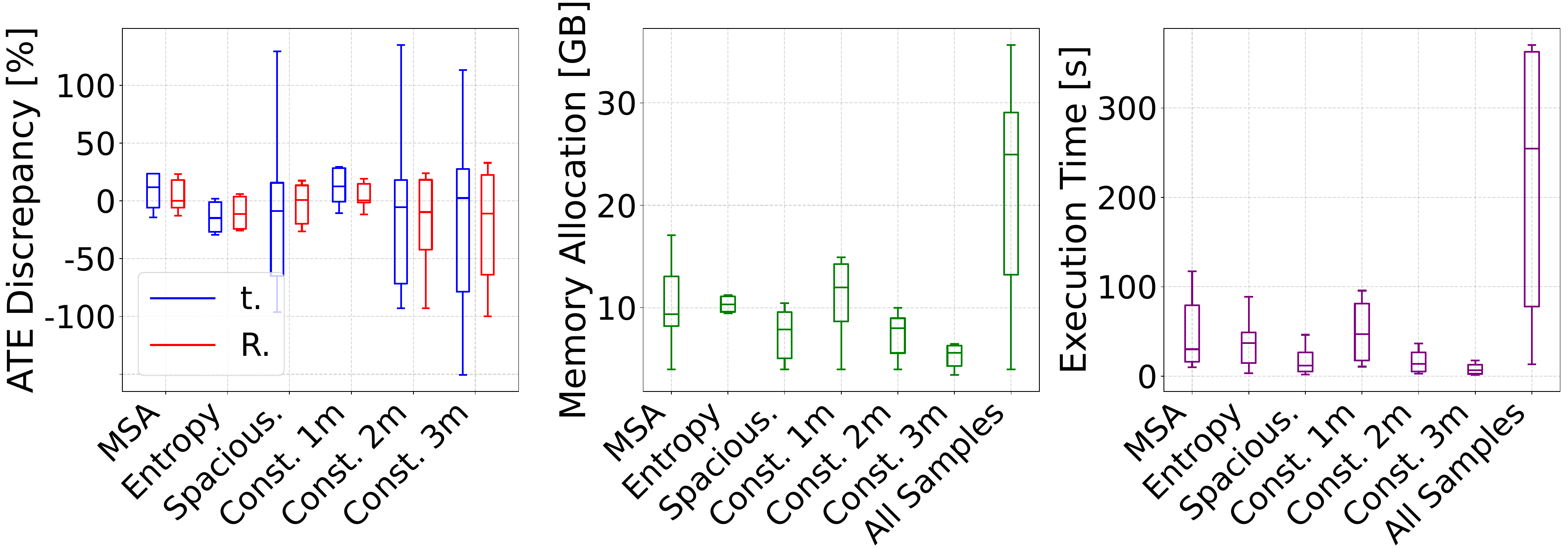}
    \caption{\textbf{Box plot comparisons.} Translational (t) and Rotational (R) Absolute Trajectory Error (ATE) difference of every method compared to the baseline of All Samples, as well as the Memory Allocation in Gigabytes (GB) and the total Execution Time in seconds (s).}
    \label{fig:deviations}
    \vspace{-0.6cm}
\end{figure}

\subsection{Quantitative and Qualitative Analysis}
For the batch experiments in Table~\ref{tab:global_evaluation}, we evaluate our approach on three datasets: \texttt{KITTI Odometry}~\cite{Geiger2012CVPR}, \texttt{MulRan}~\cite{Gisep2020mulran}, and \texttt{Apollo-SouthBay}~\cite{Weixin2019apollo}, representing diverse environments, including urban areas, rural landscapes, and complex structures like bridges and tunnels. Keyframes are sampled for all methods, and the pose graph is constructed using \texttt{GTSAM}~\cite{gtsam} with odometry provided from \texttt{KISS-ICP}~\cite{vizzo2023ral}. Loop closure candidates are identified using \texttt{OT} or \texttt{SC} descriptors with a similarity threshold of 0.8. Candidates within a 3-meter radius of the ground truth are classified as true positives and processed with \texttt{small\_gicp}~\cite{small_gicp} to estimate relative transformations. Matches with registration residuals below 0.3 meters are verified and added as edges to the pose graph, which is optimized using the Levenberg-Marquardt algorithm.

\begin{figure}[b!]
    \centering 
    \setlength{\abovecaptionskip}{-0.15cm}
    \includegraphics[width=1.0\columnwidth]{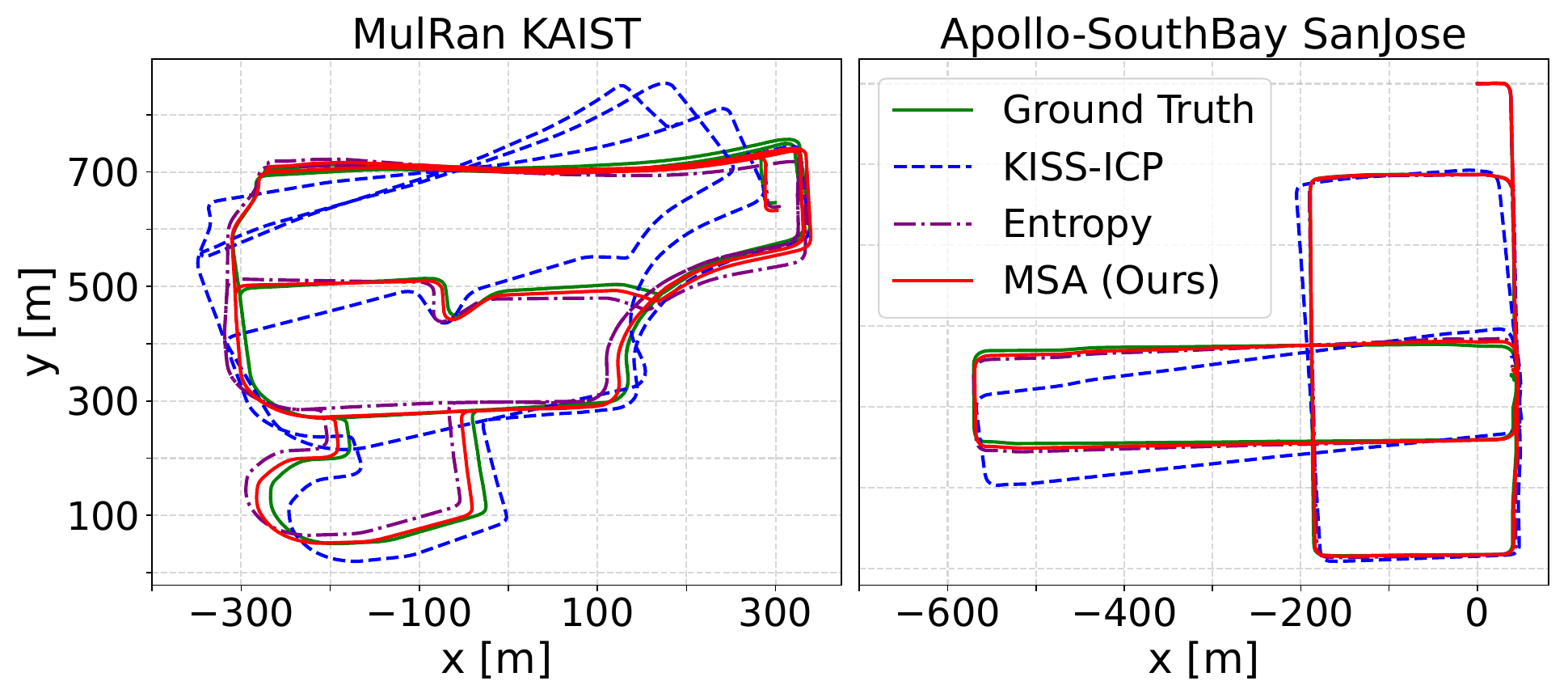}
    \caption{\textbf{Trajectory comparisons.} Comparison of the ground truth, raw \texttt{KISS-ICP}, and the sampled poses after the pose graph optimization for the proposed Minimal Subset Approach (MSA) and the entropy-based approach on the \texttt{KAIST} sequence of the \texttt{MulRan} dataset and the \texttt{SanJoseDowntown} sequence of the \texttt{Apollo-SouthBay} dataset.}
    \label{fig:trajectories}
\end{figure}

Table~\ref{tab:global_evaluation} summarizes the results for all datasets, descriptors, and methods. The metrics include the Absolute Trajectory Error (ATE) for both translation and rotation, presented as the percent improvement in the \texttt{KISS-ICP} trajectory after pose graph optimization with each method. 
Additionally, we report the Area Under the Precision-Recall Curve (PR-AUC), as we demonstrate that sampling impacts place recognition performance.
The results demonstrate that the proposed MSA consistently delivers the best performance, while constant intervals yield varying results across different environments, highlighting the absence of a universal fixed sampling interval.
In Fig.~\ref{fig:deviations}, the data from Table~\ref{tab:global_evaluation} are presented as box plots, illustrating the deviation of each method's ATE performance compared to using all samples, alongside variations in memory usage and total pipeline execution time. The proposed approach sustains overall improvements with minimal performance loss in both translation and rotation, while significantly reducing memory usage and execution time. Although other methods also reduce memory and processing overhead, their performance lacks robustness and varies considerably. Finally, Figs.~\ref{fig:concept} and~\ref{fig:trajectories} compare ground truth poses, raw \texttt{KISS-ICP} poses, and corrected trajectories for MSA and the entropy-based method post optimization. Across all illustrated trajectories, the proposed approach achieves consistently lower translation and rotation RPE than other methods, with loop closure segments closely aligning with the ground truth, demonstrating its ability to retain essential edges.

\subsection{Larger-Scale Evaluations}
\begin{table}[!b]
\centering
\caption{Results for the Larger-Scale Evaluations and the Ablation Study for the proposed approach}
\label{tab:online_evaluation}
\resizebox{\linewidth}{!}{%
\begin{tabular}{c||l||c|c|c|c|c} 
\toprule
& \textbf{Metrics} & \textbf{t. ATE}$^*$ & \textbf{R. ATE}$^*$ & \textbf{F1-MAX} & \textbf{MEM.} & \textbf{TIME}\\ 
\midrule
\multirow{11}{*}{\rotatebox{90}{\textbf{\texttt{Apollo-SB Sunnyvale}}}} & All Samples & nan & nan & nan & >128GB & nan \\ 
& {Constant 1.0m} & \textbf{98.5\%} & \textbf{98.4\%} & \underline{93.3\%} & 90.9GB & 7340s \\ 
& Constant 2.0m & 98.1\% & 97.3\% & 62.1\% & 47.2GB & 2264s \\ 
& Constant 3.0m & 97.9\% & 97.0\% & 46.3\% & 35.4GB & 1515s \\ 
& Entropy-based & 95.5\% & 93.4\% & 81.9\% & 51.0GB & 4945s \\ 
& Spaciousness & 94.9\% & 97.5\% & 46.1\% & 38.3GB & 1752s \\ 
& MSA (default) & \underline{98.2\%} & \underline{97.5\%} &\textbf{94.4\%} & 40.3GB & 1923s \\ 
\cmidrule{2-7}
& MSA (w/o $\pi_\tau$) & 94.8\% & 93.2\% & 90.1\% & 32.4GB & 1473s \\
& MSA (w/o $\rho_\tau$) & {98.9\%} & 97.7\% & 94.8\% & 43.9GB & 2124s \\
& MSA ($\alpha=10$) & 98.6\% & 97.7\% & 94.5\% & 49.2GB & 2889s \\
& MSA ($\beta=10$) & 96.1\% & 94.7\% & 90.5\% & 34.9GB & 1572s \\
\midrule
\multirow{11}{*}{\rotatebox{90}{\textbf{\texttt{Newer-College Park}}}} & All Samples & 93.9\% & 78.9\% & 99.8\% & 65.8GB & 6742s \\ 
& {Constant 0.1m} & \underline{89.7\%} & \underline{71.4\%} & \textbf{99.8\%} & 63.2GB & 4987s \\ 
& Constant 0.3m & 79.9\% & 58.5\% & 94.8\% & 28.9GB & 1262s \\ 
& Constant 1.0m & 60.7\% & 43.1\% & 64.5\% & 09.2GB & 165s \\ 
& Entropy-based & 87.3\% & 65.8\% & 92.6\% & 26.3GB & 1401s \\ 
& Spaciousness & 72.5\% & 52.5\% & 89.0\% & 27.2GB & 1096s \\ 
& MSA (default) & \textbf{92.9\%} & \textbf{73.1\%} & \underline{96.9\%} & 40.7GB & 1925s \\ 
\cmidrule{2-7}
& MSA (w/o $\pi_\tau$) & 85.9\% & 69.4\% & 93.7\% & 32.2GB & 1482s \\
& MSA (w/o $\rho_\tau$) & 93.2\% & 73.1\% & 97.8\% & 44.0GB & 2191s \\
& MSA ($\alpha=10$) & 93.7\% & 75.5\% & 98.2\% & 54.8GB & 2352s \\
& MSA ($\beta=10$) & 88.3\% & 65.9\% & 89.5\% & 35.1GB & 1749s \\
\bottomrule
\multicolumn{7}{c}{\vspace{-0.25cm}} \\ 
\multicolumn{7}{c}{$^{*}$The ATEs are presented as the percent improv. in the trajectory after PGO.}
\end{tabular}
}
\end{table}
To support our last contribution, 
that MSA achieves higher localization and place recognition performance than other adaptive methods, while reducing memory usage and computation time relative to the baseline,
we conduct more realistic, larger-scale, online experiments. Poses are processed sequentially, simulating real-time operation. Keyframes are sampled online using the previously mentioned methods, appended to the pose graph, and optimized at each step with \texttt{iSAM2}~\cite{Kaess2011isam2}. Simultaneously, \texttt{OT} descriptors are queried against past keyframes to detect loop closures, which are verified with \texttt{small\_gicp} and integrated into the graph.

\subsubsection{\texttt{Urban Driving Scenario}} 
\begin{figure*}[!t]
    \centering
    \subfloat[\begin{tabular}{c}
         \texttt{\textbf{Apollo-SouthBay Sunnyvale}}
    \end{tabular}]{%
        \includegraphics[width=0.98\columnwidth]{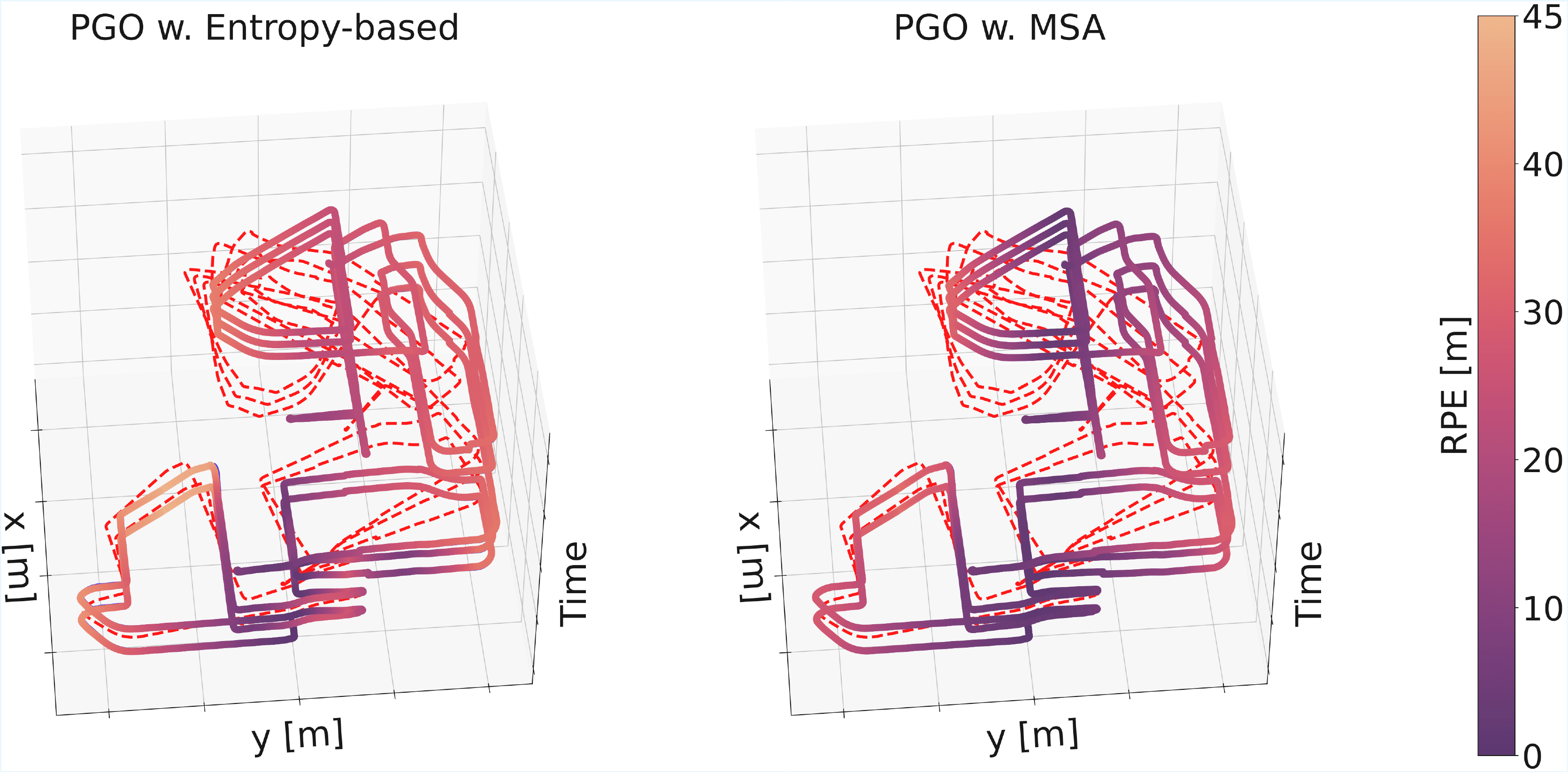}}
    \vspace{-0.1cm}
    \hfill
    \subfloat[\begin{tabular}{c}
         \texttt{\textbf{Newer-College Park}}
    \end{tabular}]{%
            \includegraphics[width=1.0\columnwidth]{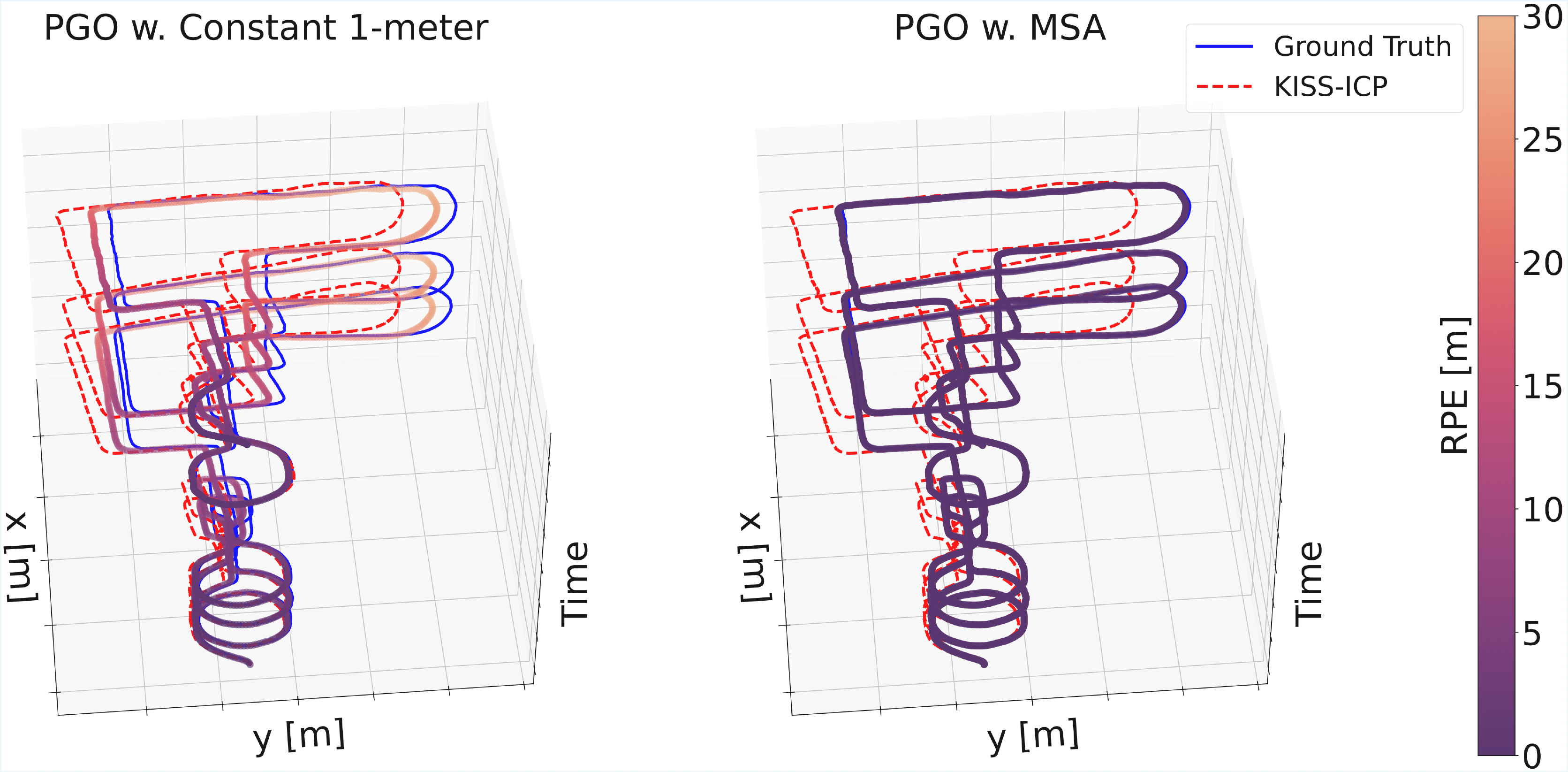}}
    \caption{\textbf{Trajectory comparisons.} The ground truth and raw \texttt{KISS-ICP} poses compared against the sampled and optimized poses using the proposed Minimal Subset Approach (MSA) and the entropy-based method for the \texttt{Sunnyvale} and the constant 1-meter approach for the \texttt{Park}. 
    The color gradient represents the translational Relative Pose Error (RPE). Grid spacing corresponds to 0.5x0.5 km$^2$ for the left figure and 50x50 m$^2$ for the right figure.}
    \label{fig:trajectories_2}
    \vspace{-0.5cm}
\end{figure*}

For this scenario we use one of the datasets from the batch experiments, \texttt{Apollo-SouthBay}, focusing on the \texttt{SunnyvaleBig-} \texttt{loop} sequence, which spans over 100 kilometers.
Table~\ref{tab:online_evaluation} reports the translation and rotation ATE improvements, F1-max score, total allocated memory, and processing time. Notably, when using all samples, the system exhausts memory, approximately 70\% into the sequence, preventing the calculation of the metrics. The other sampling methods reduce memory usage while maintaining excellent performance. This is likely due to the sequence revisiting the same locations multiple times, enabling frequent loop closure edges that mitigate accumulated drift.
In Fig.~\ref{fig:trajectories_2}, the optimized trajectory of the entropy-based method is compared to MSA. The visualization highlights MSA's better performance, achieved with lower memory allocation and a higher F1-max score. Lastly, Fig.~\ref{fig:memory} shows the memory usage over time, illustrating how sampling methods scale more efficiently compared to retaining all samples or using a constant 1-meter interval.

\subsubsection{\texttt{Campus Scenario}} To provide additional insights, we evaluate a different dataset, the \texttt{Newer-College}~\cite{Ramezani2020newercollege}, and more specifically the \texttt{Park} sequence, a 30-minute mapping session with multiple loops. This scenario, resembling a mobile ground robot, differs significantly from \texttt{Apollo-SouthBay} due to its more confined environment and denser frame sampling, approximately 0.1 meters compared to 1 meter. These differences highlight how distinct missions impose unique requirements.

\begin{figure}[!b]
    \centering
    \setlength{\abovecaptionskip}{-2pt}
    \includegraphics[width=0.95\columnwidth]{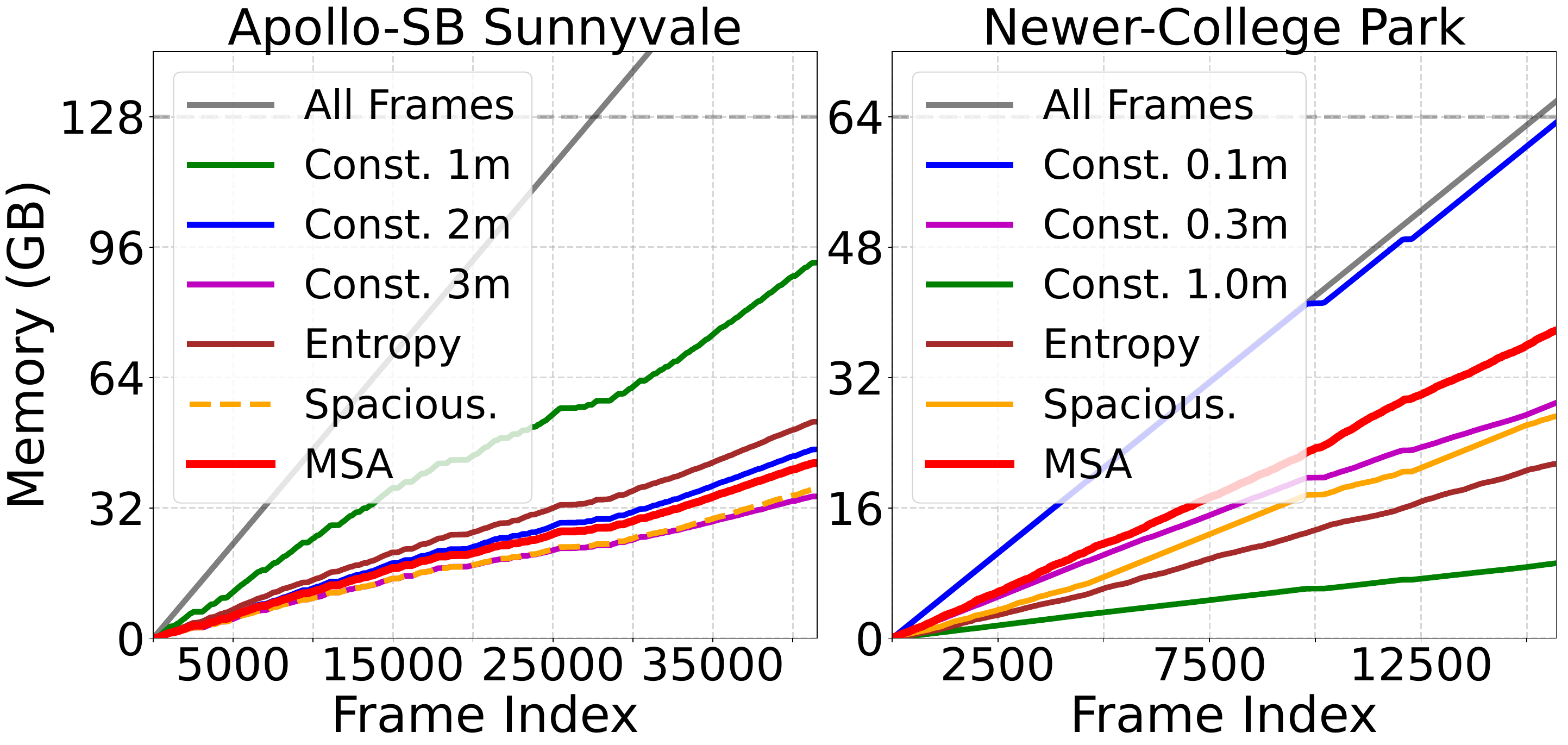}
    \caption{\textbf{Memory allocation.} Memory against each acquired frame for the different sampling methods. Retaining all samples exceeds system memory of 128GB, demonstrating the need for efficient and scalable algorithms.}
    \label{fig:memory}
    \vspace{-0.2cm}
\end{figure}

\begin{figure}[!b]
    \centering
    \setlength{\abovecaptionskip}{-2pt}
    \includegraphics[width=0.95\columnwidth]{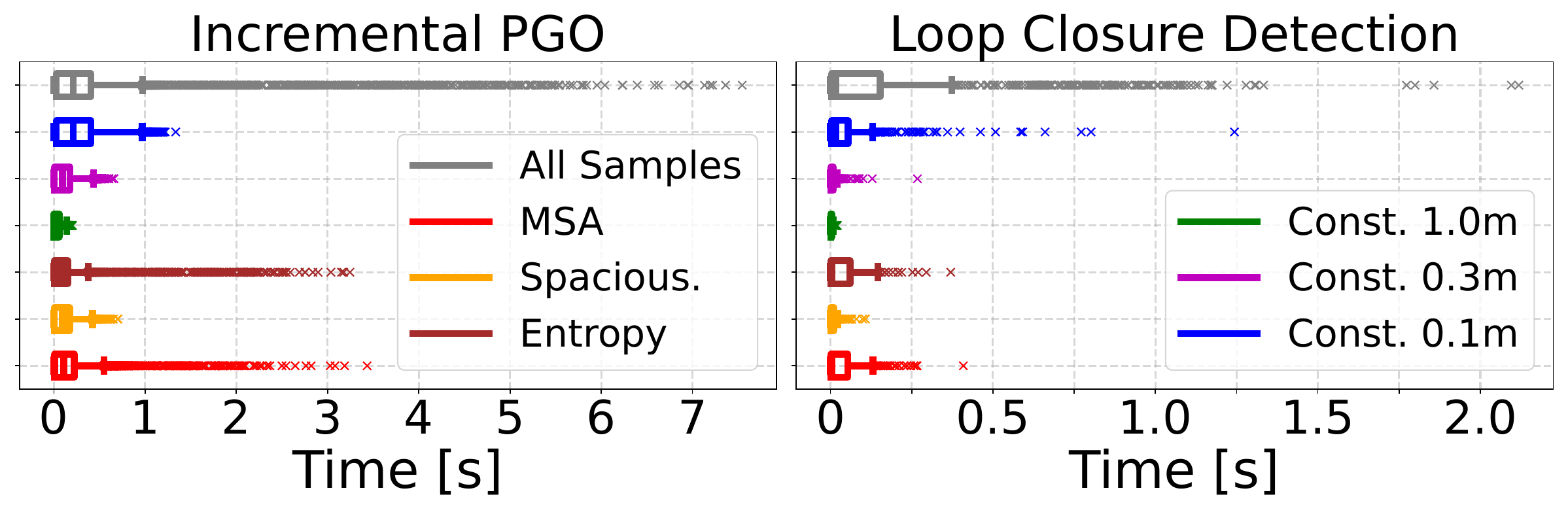}
    \caption{\textbf{Processing times.} The distributions of the incremental pose graph optimization convergence (left), and the loop closure detection (right) time, per new frame, for the \texttt{Park} sequence of the \texttt{Newer-College} dataset.}
    \label{fig:times}
\end{figure}

From Table~\ref{tab:online_evaluation}, constant intervals show that, a 1-meter sampling, severely underperforms in this scenario, retaining only 7\% of the frames. This leads to significant performance losses in both ATE and FPR. To address this, we include lower constant intervals of 0.1 and 0.3 meters for comparison. While these improve performance, our proposed MSA outperforms them by prioritizing informative sampling, providing more valuable nodes to the pose graph. The entropy-based method closely follows, emphasizing the importance of spatial features in sampling. Here, we note that both the spaciousness- and entropy-based methods required additional tuning for this scenario.
As in the previous experiment, Fig.~\ref{fig:trajectories_2} compares the optimized trajectory of MSA with the 1-meter approach, which highlights MSA's higher metric localization performance.
Figs~\ref{fig:memory} and~\ref{fig:times} further compare memory allocation over time and processing times. Fig.~\ref{fig:times} presents box plots for incremental PGO (left) and Loop Closure Detection (right) processing times per step. Notably, both tasks exhibit significantly high processing times, potentially causing delays in real-world missions. 
While lower fixed intervals (0.1 and 0.3 meters) reduce processing time, 0.3-meter and 1-meter intervals show subpar performance, and the 0.1-meter interval demands substantial memory allocation since it is very close to the native sampling of the dataset.
Interestingly, methods achieving good ATE performance, such as the entropy-based and MSA, demonstrate higher outliers in the incremental PGO, suggesting that valuable edges are added to the pose graph, contributing to their better performance.

\subsection{Ablation Study} \label{subsec:ablation}
\begin{figure}[!t]
    \centering
    \setlength{\abovecaptionskip}{-2pt}
    \includegraphics[width=1.0\columnwidth]{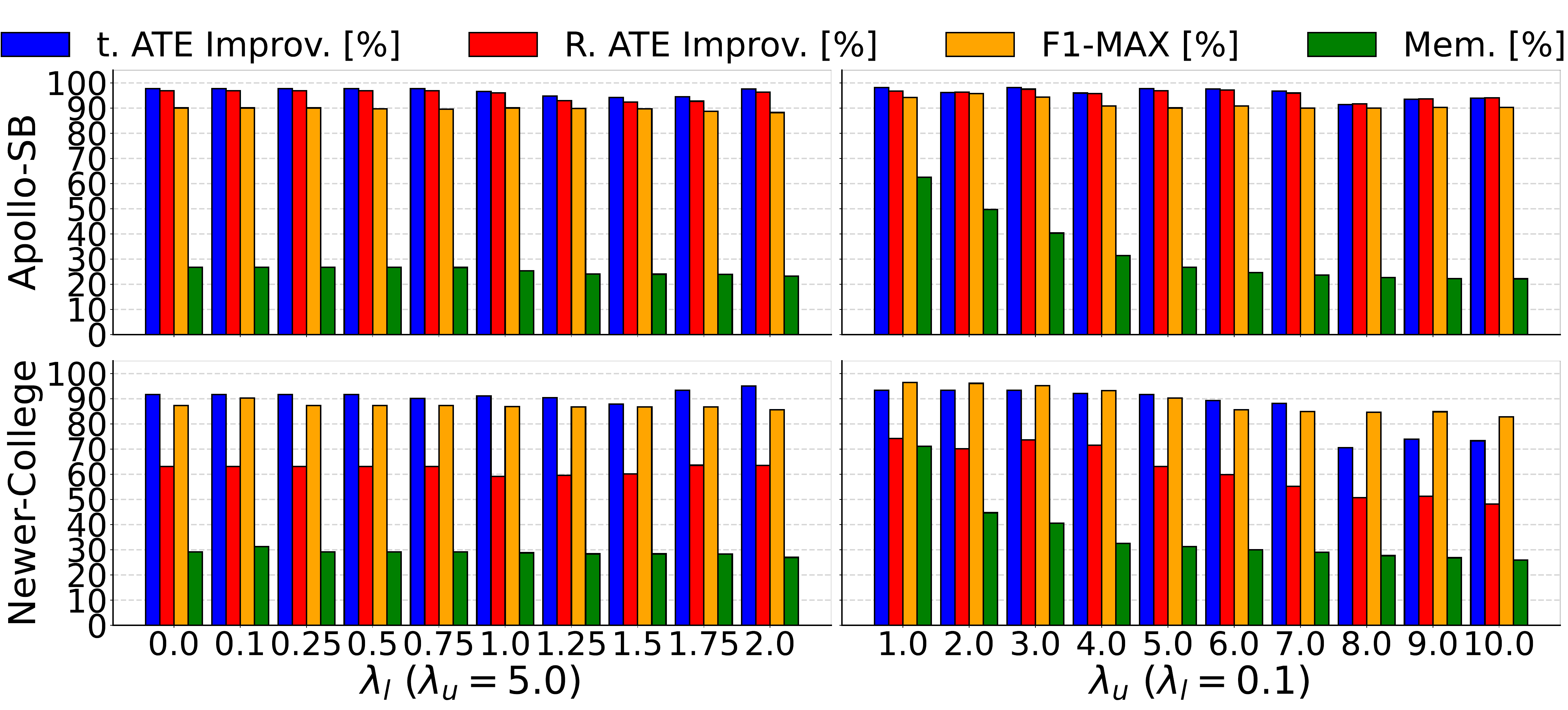}
    \caption{\textbf{Adaptive bounds ablation.} For each dataset, we sweep $\lambda_l$, while keeping $\lambda_u = 5.0$, and $\lambda_u$ while keeping $\lambda_l = 0.1$. We report translational and rotational ATE improvement, the F1–max from the precision and recall curves, and the memory allocation compared to all samples.}
    \label{fig:ablation_distance}
    \vspace{-0.5cm}
\end{figure}

In all experiments, we maintained constant hyperparameters ($\alpha = \beta = 1$ and $\lambda_l=0.1$, $\lambda_u=3$), demonstrating that the proposed method adapts seamlessly to different environments and descriptors without requiring manual tuning. To further analyze the impact of each optimization term, we conducted an ablation study, with results shown in Table~\ref{tab:online_evaluation}. Specifically, we tested scenarios where only one term, $\rho_\tau$ or $\pi_\tau$, was used, and scenarios where the weight of one term was increased to 10 while keeping the other at 1. The results indicate that emphasizing the \textit{information preservation} term or using it exclusively improves performance but leading to increased number of keyframes sampled. Conversely, prioritizing \textit{redundancy minimization} significantly reduces memory usage but negatively impacts performance.
Based on these, practitioners can decide whether to prioritize memory or performance, depending on their specific application requirements.
Additionally, Fig.~\ref{fig:ablation_distance} reports an ablation of the distance bounds, sweeping $\lambda_l\!\in[0,2]$ and $\lambda_u\!\in[1,10]$, and evaluating all previously discussed metrics. Accuracy remains largely stable, while memory is more sensitive: larger $\lambda_u$ reduces memory on \texttt{Apollo} with minimal loss but degrades accuracy on \texttt{Newer–College}. The latter is a more confined environment where salient features are not widely spaced, which induces rapid viewpoint changes; as $\lambda_u$ grows, consecutive frames become too far apart for descriptors to remain smooth, weakening the information-preservation term and reducing the reliability of PGO edges. We therefore adopt $\lambda_l=0.1$ to avoid near-duplicates and $\lambda_u=3$ to preserve descriptor and odometry locality.

\section{Discussion and Conclusions}
The results highlight that no universal constant sampling interval is suitable for all scenarios. While a constant 1-meter interval performs well in urban driving datasets like \texttt{Apollo-SouthBay}, it performed poorly in the mobile robotic scenario of \texttt{Newer-College}. This reinforces our core argument: adaptive methods that account for both place recognition and pose graph optimization are essential. Naive heuristic approaches based on spaciousness or entropy require extensive tuning and prior environmental knowledge, limiting applicability.
MSA consistently performs well across scenarios without parameter tuning, and is effective with both widely used hand-crafted and learning-based descriptors. As future work, we plan to integrate MSA with full global localization pipelines such as \texttt{LCDNet}~\cite{cattaneo_lcdnet_2022} and \texttt{RING++}~\cite{Xu2023ring}.

In conclusion, as robotics scales to larger and longer missions, efficient and scalable sampling becomes crucial. The Minimal Subset Approach reduces redundant samples while maintaining robust performance. Across multiple datasets, MSA delivers better place recognition and superior ATE/RPE than adaptive methods, and compared to using all samples it reduces memory consumption and computation time.







\bibliographystyle{./IEEEtranBST/IEEEtran}
\bibliography{./IEEEtranBST/IEEEabrv,root}


\end{document}